\documentclass[sigconf]{acmart}
\usepackage{array, multirow, bigdelim, makecell, booktabs,amsmath,tcolorbox} 
\usepackage{multirow,balance}
\usepackage{booktabs}
\usepackage{caption}
\usepackage[ruled,vlined]{algorithm2e}
\copyrightyear{2021}
\acmYear{2021}
\setcopyright{rightsretained}
\acmConference[AIES '21]{Proceedings of the 2021 AAAI/ACM Conference
on AI, Ethics, and Society}{May 19--21, 2021}{Virtual Event, USA}
\acmBooktitle{Proceedings of the 2021 AAAI/ACM Conference on AI,
Ethics, and Society (AIES '21), May 19--21, 2021, Virtual Event,
USA}\acmDOI{10.1145/3461702.3462576}
\acmISBN{978-1-4503-8473-5/21/05}

\begin{document}
\fancyhead{}

\title{More Similar Values, More Trust? - the Effect of Value Similarity on Trust in Human-Agent Interaction}


\author{Siddharth Mehrotra}
\affiliation{%
  \institution{Delft University of Technology}
  \streetaddress{Van Mourik Broekmanweg 6}
  \city{Delft}
  \country{The Netherlands}
  \postcode{2628 XE}
}\email{s.mehrotra@tudelft.nl}

\author{Catholijn M. Jonker}
\affiliation{%
  \institution{Delft University of Technology \& LIACS, Leiden University}
  \streetaddress{Van Mourik Broekmanweg 6}
  \city{Delft}
  \country{The Netherlands}
  \postcode{2628 XE}}
\email{c.m.jonker@tudelft.nl}

\author{Myrthe L. Tielman}
\affiliation{%
  \institution{Delft University of Technology}
  \streetaddress{Van Mourik Broekmanweg 6}
  \city{Delft}
  \country{The Netherlands}
  \postcode{2628 XE}
}\email{m.l.tielman@tudelft.nl}

\renewcommand{\shortauthors}{Trovato and Tobin, et al.}

\begin{abstract}
  As AI systems are increasingly involved in decision making, it also becomes important that they elicit appropriate levels of trust from their users. To achieve this, it is first important to understand which factors influence trust in AI. We identify that a research gap exists regarding the role of personal values in trust in AI. Therefore, this paper studies how human and agent Value Similarity (VS) influences a human's trust in that agent. To explore this, 89 participants teamed up with five different agents, which were designed with varying levels of value similarity to that of the participants. In a within-subjects, scenario-based experiment, agents gave suggestions on what to do when entering the building to save a hostage. We analyzed the agent's scores on subjective value similarity, trust and qualitative data from open-ended questions. Our results show that agents rated as having more similar values also scored higher on trust, indicating a positive effect between the two. With this result, we add to the existing understanding of human-agent trust by providing insight into the role of value-similarity.
\end{abstract}




\maketitle

\section{Introduction}
In the Indian epic Mahabharata, Arjun and Bhima are important characters. They go through common struggles and trust in each other's abilities. 
They challenged Jarasandha and Chitrasena's (\textit{two kings}) armies and fought for Kampilya together (\textit{a capital kingdom}). What made them have so much trust in each other? According to Rajagopalachari \cite{rajagopalachari1970mahabharata}, the most compelling reason 
was that they shared similar values. In this paper, we explore how we can take inspiration from this story when trying to understand trust in AI.


As AI systems gain complexity and become more pervasive, it becomes crucial for them 
to elicit appropriate trust from humans. We should avoid under-trust, as it would mean not making optimal use of AI. Yet we should also avoid over trust, as relying on AI systems too much could have serious consequences \cite{parasuraman1997humans}. As a first step towards eliciting appropriate trust, we need to understand what factors influence trust in AI agents. Despite the growing attention in research on trust in AI agents, a lot is still unknown about people's perceptions of trust in AI agents \cite{glikson2020human}. Therefore, we wish to know what it is that makes people trust or distrust AI? In this paper, we 
see trust as multi-dimensional as suggested by Roff and Danks \cite{roff2018trust}. On the one hand, trust corresponds to reliability and/or predictability and on the other hand trust depends upon people’s values, preferences, expectations, constraints, and beliefs. Various studies have examined how trust is attributed according to the first dimension \cite{ryan2020ai, chavaillaz2016system}, but fewer have investigated the second dimension, where the focus is on people's shared values \cite{cranefield2017no}. The implication of the latter dimension for the design of agents is on how to design these agents with respect to values as different people prioritize different values, which in turn guides how people behave and judge the behavior of others 
\cite{friedman2008value}. 

We argue that there is a research gap in understanding the role of values on the trust a human has in that agent. 
Siegrist et al. state \cite{siegrist2000salient}: 
\begin{tcolorbox}[left=.5em,right=.5em,boxrule=0pt]
	 ``\textit{people base their trust judgments on whether they feel that the agency shares similar goals, thoughts, values, and opinions''}
	\end{tcolorbox}
For example, if you value \emph{cost-efficiency} over \emph{aesthetics} when it comes to buildings, you would probably trust an architect more if they have shown that \emph{cost-efficiency} is also important to them. 
Regarding trust in AI systems, we resonate with Tolmeijer et al. \cite{tolmeijer2020taxonomy} in observing the potential for overlap and contrast with the psychology, ethics, and pragmatics of trust between humans. 
Based on this, we hypothesize that the trust of humans in AI agents is positively correlated to the similarity of the values of those agents and humans.
Taking this approach forward in AI agent research, we examine the effect of (dis)-similarity (of human \& agent's values) on a human's trust in that agent. We design five different agents with varying value profiles so that for any human, some of these are more similar and some less similar to the value profile of that human. The agents team up with participants for a risk-taking task scenario for which they have to interact and decide on the appropriate action to take. Participants evaluate the agents based on how much they trust each agent and their perceived Value Similarity (VS).  

In the remainder of this paper, we first review related work on value similarity and give an overview of existing literature on the use of values to promote trust. We then describe the design of the agents we use in the experiments, and the setup of our user study. We discuss our results and conclude with potential applications and limitations of our work.

\section{Related Work}
Trust within the AI domain has been explored mostly in contexts such as decision making \cite{shvo2019towards}, examining/assessing user's trust \cite{ogawa2019humans}, and improving the system performance \cite{partovi2019relationship}. We argue that it is important to also consider the similarity of personal values when researching trust. But can an AI agent have personal values? Increasingly, researchers are trying to incorporate values in AI systems, especially systems which are in some way involved in (helping humans with) decision making. 

Winikoff argue value-based reasoning to be an essential prerequisite for having appropriate human trust in autonomous systems \cite{winikoff2017towards}. This thought echoes with prior work by Banavar \cite{guru_2016}, van Riemsdijk et al. \cite{van2015creating} and Mercuur et al. \cite{mercuur2019value}. More recently, Cohen et al. acknowledge \cite{cohen2019trusted}: 
\begin{tcolorbox}[left=.5em,right=.5em,boxrule=0pt]
	 ``\textit{Human users will be disappointed if the AI system makes no effort to represent or reason about inherent social values that users would like to see reflected.''}
	\end{tcolorbox}
Most practical work on implementing human values in AI system focuses on plan selection \cite{cranefield2017no}, user-agent value alignment \cite{shapiro2002user} and studying agent's value driven behaviour \cite{dechesne2013no}. One of the earlier attempts to look at the effect of similarity of values on trust was made within social science research by Siegrist et al. \cite{siegrist2000salient}. They showed similar values, and trust depends upon each other in human-human interaction. Their findings resonated with Sitkin and Roth \cite{sitkin1993explaining} who report that interpersonal trust is based on shared values. On these lines, Vaske et al. showed that as salient value similarity increases, social trust in the agency increases \cite{vaske2007salient}. Their findings showed how understanding the value similarity between Colorado residents and United States department of agriculture, resulted in social trust and attitudes towards wildland fire management.

Recently, researchers have been interested in using this concept of value similarity for AI systems as well. Cruciani et al. designed an agent based model showing how similarity in values can be a successful driver for cooperation in the regulation and design of public policies \cite{cruciani2017dynamic}. They analyze their simulation experiment by looking at how and, how much agents cooperate with similar others. The key takeaway message is the introduction of value similarity for investigating what ultimately motivates trust-building processes. However, their work used predetermined memory coefficient for simulation agents to study coordination and was not validated with human participants. Additionally, Chhogyal and colleagues designed a formal trust assessment model \cite{ijcai2019-28}. In their work, they developed value-based trust assessment functions and showed how they lead to trust sequences. However, they did not consider value preferences and neither validated the model with human participants. Building on these works, our research is looking for a deeper understanding regarding the effect of value similarity on trust in a risk taking scenario accounting for the perception of human participants instead of providing simulation based results. 

\section{Method}
The primary goal of our study is to understand how (perceived) value similarity affects trust. We focused on exploring how users' trust is affected by interaction with different agents with varying value similarity. More specifically, we have the following hypothesis:
\begin{tcolorbox}[left=.5em,right=.5em,boxrule=0pt]
\textbf{Value similarity} between the user and the agent \textbf{positively} affects the trust a user has in that agent.
\end{tcolorbox}
\subsection{Creation of value profiles}
We used the Schwartz Portrait Value Questionnaire (PVQ) \cite{schwartz2012overview} to draw each participant's user profiles which consist of ten value dimensions. 
There are statements about each value dimension in the PVQ. 
Participants were asked to read carefully and respond to how each statement resonate with them as a person on a scale of 1-6, where ‘1’ means ‘\textit{very much like me}’ and ‘6’ implies ‘\textit{not at all like me}'.

\begin{figure}[h]
\centering
\includegraphics[width=0.3\textwidth]{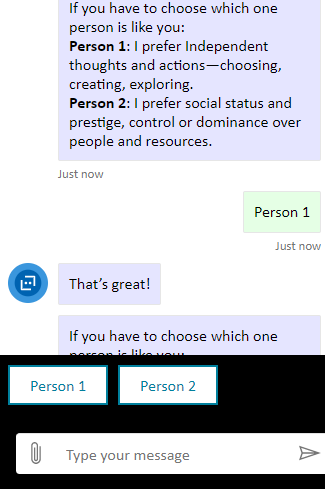}
\caption{Human-AI agent interaction chatbot testbed with HTML front-end.}
\label{fig:chatbot}
\end{figure}

For each ‘\textit{very much like me}’ we assigned a score of 1 and for each ‘\textit{not at all like me}' a score of 6 to that value. Furthermore, we created an actual value profile for each user based on their rank\footnote{We define rank as a position in the hierarchy of importance of the values.} (\textit{refer column `PVQ Score' in table 1}). 
We combined the first two values according to rank as group one, the second two values as group two, and so on till group five. We grouped ten values into five groups with two values each. Sometimes, a group can have more than two values because multiple values could receive the same final score. To resolve this conflict, we employ Algorithm 1 (\textit{see Appendix 1}) to get user priority.
\begin{table}[t]
\centering
\small
\captionsetup{font=small}
\centering
\setlength{\tabcolsep}{3pt}
\renewcommand{\arraystretch}{1.65}
  \caption{An example of generating value profiles of agents based on human value profile. \textit{Rank} represents order of the values, \textit{PVQ} represents the PVQ scores by participants, \textit{Corrected} represents scores after applying the algorithm 1. Lower scores corresponds to higher ranks. C1 showcases conflict between three values for group two. Group 1 (G1) - Group 5 (G5) are groups for the first two ranks, the second two ranks, and so on... representing five different agents. }
\label{tab:locations}
  \begin{tabular}{llll}\toprule
    \textbf{\textit{Rank}} & \textbf{\textit{PVQ Score}} & 
    \textbf{\textit{Corrected}} & \textbf{\textit{Value}} \\ \midrule
    1 & 1 & 1.0 & Security\hspace{2.5em}\rdelim\}{2}{*}[\textit{Group 1}: G1] \\
    2 & 1 & 1.0 &Self Direction  \\
    3 &  2 \hspace{0.5em}\ldelim\}{3}{*}[] &\hspace{-2.0em}\ldelim\{{3}{*}[C1]1.90 & Traditional\hspace{1.35em}\rdelim\}{2}{*}[\textit{Group 2}: G2]\\
    4 & 2  &1.95 & Conformity\\
    5 & 2  &2.0& Universalism\hspace{0.5em}\rdelim\}{2}{*}[\textit{Group 3}: G3]\\  
    6 & 3  & 3.0 & Power\\
    7 & 4 & 4.0 & Benevolence\hspace{0.7em}\rdelim\}{2}{*}[\textit{Group 4}: G4]\\
    8 & 4  & 4.0 & Hedonism\\
    9 & 5  & 5.0 & Achievement\hspace{0.6em}\rdelim\}{2}{*}[\textit{Group 5}: G5] \\
    10 & 6  & 6.0 & Stimulation\\\bottomrule
  \end{tabular}
\end{table}
For example, in table 1, there are three values with a score of 0.9 (\textit{refer set C1}); and we needed only two values for each group. Therefore, participants were asked to choose one value over another based on the meaning of two values (\textit{refer Figure \ref{fig:chatbot}}) following algorithm 1. 
In our user-study, we did not come across a conflict case where there were more than four values with the same PVQ score. 

\subsection{Agents and the scenario}
We designed a “save a hostage game” in which each participant interacts with five different agents that provided tips and suggestions to save the hostage. The task was inspired by prior work from Wang et al. \cite{wang2016impact}. In our game, agents were featured with varying value profiles.
\subsubsection{Agents and Value Similarity:}
For each participant, we created five different agents with descending value similarity profiles from G1 to G5 (\textit{see table 1 for example}). 
G1 is the agent who promotes the two top ranked values of the participant, G2 agent which promotes the values ranked 3 and 4, G3 promotes the values ranked 5 and 6, etc. (so the values that each agent promotes can differ for each participant depending on their PVQ outcome). 


\subsubsection{Scenario and agent explanation:}
We provided the following scenario to our participants in which they need to team up with AI agents to rescue a hostage: ``\textit{A hostage is being held inside a building in a market place. The objective is to gather intelligence regarding the building. All five different AI agents are equipped with sensors, infrared cameras, and metal detectors. The AI agents can perform the security check of the building and inform you regarding any danger. You need to make a decision for the action to be taken based on the AI agent's advice before you enter the building}."

We used the agent's names as A, B, C, D, and E mapping to G1-G5 in our user-study. Each agent provides a suggestion to the user based on their prior common knowledge and values that are of utmost importance. A piece of prior common knowledge for all the agents was \textit{``I have searched the overall place and have found traces of the gun powder. I recommend that you take protective gear \& armor shield with you''}. 

We designed our suggestions based on the values following the notion of situation vignettes in the work by Strackand and Gennerich \cite{strack2011personal}. The values were expressed through the suggestions the agent gave. Two researchers from Computer Science background and one from Cognitive Science background brainstormed together and generated sentences that formed suggestions by the agent. Overall, three iterations of each suggestion was performed to reach the final outcome.

For example, an agent provides the following suggestion based on prior knowledge plus their values from group one - security and self-direction: \textit{``I have searched the overall place and have found traces of gun powder. I recommend that you take protective gear \& an armor shield with you. For any action you take, do follow social orders \& protocols. You should hand over the kidnapper to the police to abide by the national security laws. However, it's up to you what equipment you want to take inside the building \& how you wish to deal with the situation}.''
\subsection{Participants}
We estimated our sample size with the G-Power tool from Faul et al. \cite{faul2007g}. Our effect size was 0.30 (\textit{medium}) with linear regression as our choice for modelling variables. G-power calculated our required sample size of 81. We recruited 101 participants from the different universities' mailing list. Twelve participants could not pass our attention check, leaving 89 participants aged between 22 and 32 years old (M= 25.6; SD = 0.94). Each participant signed an informed consent form before the user-study. This study was approved by the ethics committee of our institution, ID number 1313.

We asked our participants to provide their cultural backgrounds before starting the user-study. Most of our participants were from the Europe region (34), followed by Asia Pacific (29), Americas (13), Middle East and Africa (9), and Oceania (2). Two participants did not provide their background.
\subsection{User study test bed}
We implemented an online version of our scenario to study the impact of manipulating value similarity on trust. The test bed consists of a chatbot application that can be accessed from a web browser (\textit{see figure 1}). We used Microsoft Power Apps API \footnote{https://powerapps.microsoft.com/en-us/} to generate suggestions by the agents. These were displayed on the participant's chatbot interface, which sends data back to the test bed server. The user study test bed can be found at
(website blinded for the review).
\subsection{Procedure}
Each participant first read an information sheet about the study and then fill out the background survey. Next, participants were asked to complete the PVQ to get their value profiles. After filling the PVQ, the system checked for any conflicts in value groups and asked the participant to choose one over another. Following this, the scenario was introduced to the participant.

All five agents interacted with the participant one by one. The order of appearance of the agents was randomly assigned in such a way that the order was different for each participant. Each agent appears with a small greeting and provides their suggestion. After each agent gave the suggestion, the participant was asked to fill questions from the Value Similarity Questionnaire (VSQ) \cite{siegrist2000salient} and questions from the Human-Computer Trust Scale (HCTS) \cite{gulati2019design}. In HCTS, trust is divided into three attributes, namely: general trust, benevolence, and willingness \textit{(see appendix 2 for details)}. The study was designed to be completed in 30 minutes. Participants were given a chance to participate in a raffle worth 5x20 Euro.
\section{Results}
We analyzed the results of our study, including both the subjective rating responses to the value similarity, the trust questionnaire and, the explanations provided by the participants for selecting an agent. We were primarily interested in the effect of value similarity on trust. Thus, for this paper, we focus on understanding the effect on trust by manipulating the value similarity. We call VSQ responses from participants as subjective value similarity. As part of our analysis, we first ran a Shapiro-Wilks test for normality. Since the distribution was not normal, we used non-parametric tests for our analysis.
\subsection{Manipulation check}
\begin{figure}[h]
\centering
\includegraphics[width=0.5\textwidth]{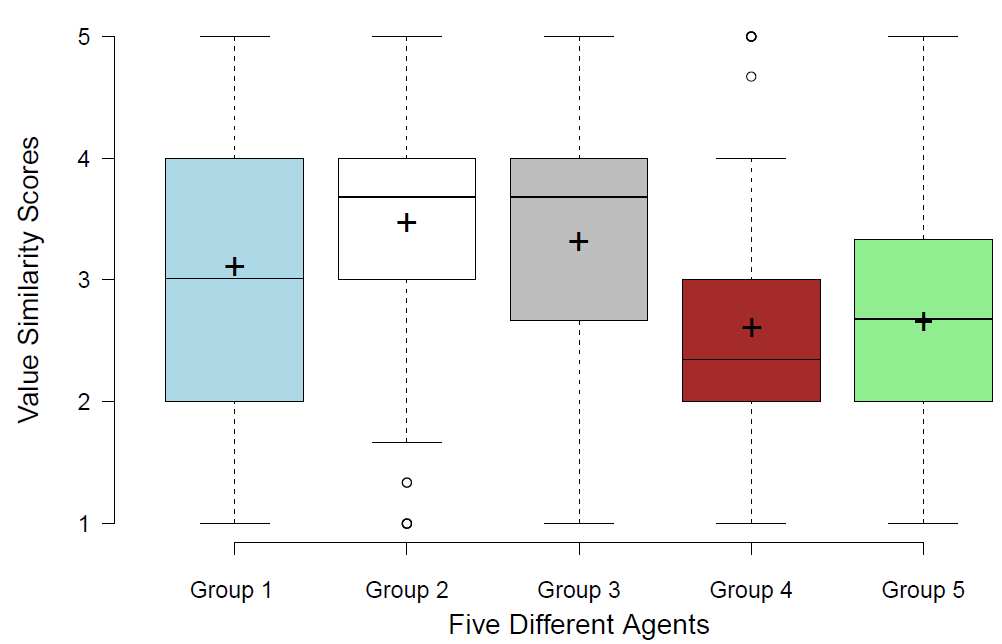}
\caption{Mean subjective VS scores for all VSQ given by participants for the five agents. The horizontal line indicates the median and the plus sign the mean value for VS scores.}
\label{fig:failedmanipulation}
\end{figure}

We tried to manipulate value similarity in this study. However, to check whether our most \textit{`similar'} to least \textit{`similar'} agent were actually perceived as most and least similar, we also measured subjective value similarity.  From figure \ref{fig:failedmanipulation}, we see that the \textit{`G2'} agent scored higher than the `\textit{G1'} agent, $\chi^2_r$ \textit{= 11.725, p $<$ .05}. This was in contradiction with the manipulation that we performed. In an ideal case, we expect the VSQ ratings to follow the order as G1 agent receives the highest VS score and G5 the least. This showcases that our manipulation did not work as expected.
Considering this, we now only focus upon value similarity as a whole rather than distribution /categorization of five agents. Therefore, in the rest of the paper we disregard our categorization of the agents.
\subsection{Correlation between Value Similarity and overall Trust}
We analyzed responses for the VSQ and HCTS to see to what extent subjective value similarity has an affect on trust. 
\begin{tcolorbox}[left=.5em,right=.5em,boxrule=0pt]
	A Kendall rank correlation test revealed that VS and trust are significantly moderately correlated in accordance with Ratner \cite{ratner2009correlation} with a correlation coefficient of \textit{0.46} and \textit{p $<$ 0.05}.
	\end{tcolorbox}

We also applied a simple linear regression model to predict a quantitative outcome of trust based on a single predictor variable \textit{i.e.} value similarity. To check linear model assumptions, we used the `GVLMA' - Global Validation of Linear Models Assumptions \cite{pena2006global} which provides a testing suite for many of the assumptions of general linear models. The four assumptions: normality, heteroscedasticity, linearity and, uncorrelatedness of the model were acceptable by the GVLMA, \textit{see appendix 3}. Linear regression showed that both the p-values for the intercept and the predictor variable were highly significant indicating a significant association between the variables, \textit{refer appendix 4}. Our goodness-of-fit measures showcase $\sigma$ = 0.984 meaning that the observed trust values deviate from the true regression line by approximately 0.984 units on average on a scale from one to five and \textit{r}\textsuperscript{\textit{2}} was 0.308. 

Finally, to seek an answer to the problem: ``can we predict trust from VS?'', we need to look at the intercept and residuals of the linear regression. On observing the intercept and residuals we have good reason to believe an overall effect of value similarity on trust.  This confirms how closely VS and trust are related. Additionally, we wished to check if differences in cultural background of participants affected the effect of VS on trust. However, because our sample size was very diverse there were not enough participants from any distinct cultural background for a statistical comparison between them. Such an effect is potentially important, but future work would need to be done to test for this. 
\subsection{Benevolence, and Willingness as attributes of overall trust}
We examined the results of HCTQ as attributes of trust namely benevolence, willingness and general trust on value similarity. We already reported the results of the general trust in previous sections. Now, we focus ourselves to Benevolence and Willingness. A Kendall tau correlation was performed to determine the relationship between benevolence, willingness and value similarity. There was a medium, positive correlation between benevolence and value similarity, which was statistically significant (\textit{r} = .47, n = 436, \textit{p} = .0002). Similarly, for willingness, correlation was found to be positive (\textit{r} = .37, n = 436, \textit{p} = .0002).


\subsection{Qualitative data analysis}
\begin{figure}[h]
\centering
\includegraphics[width=0.48\textwidth]{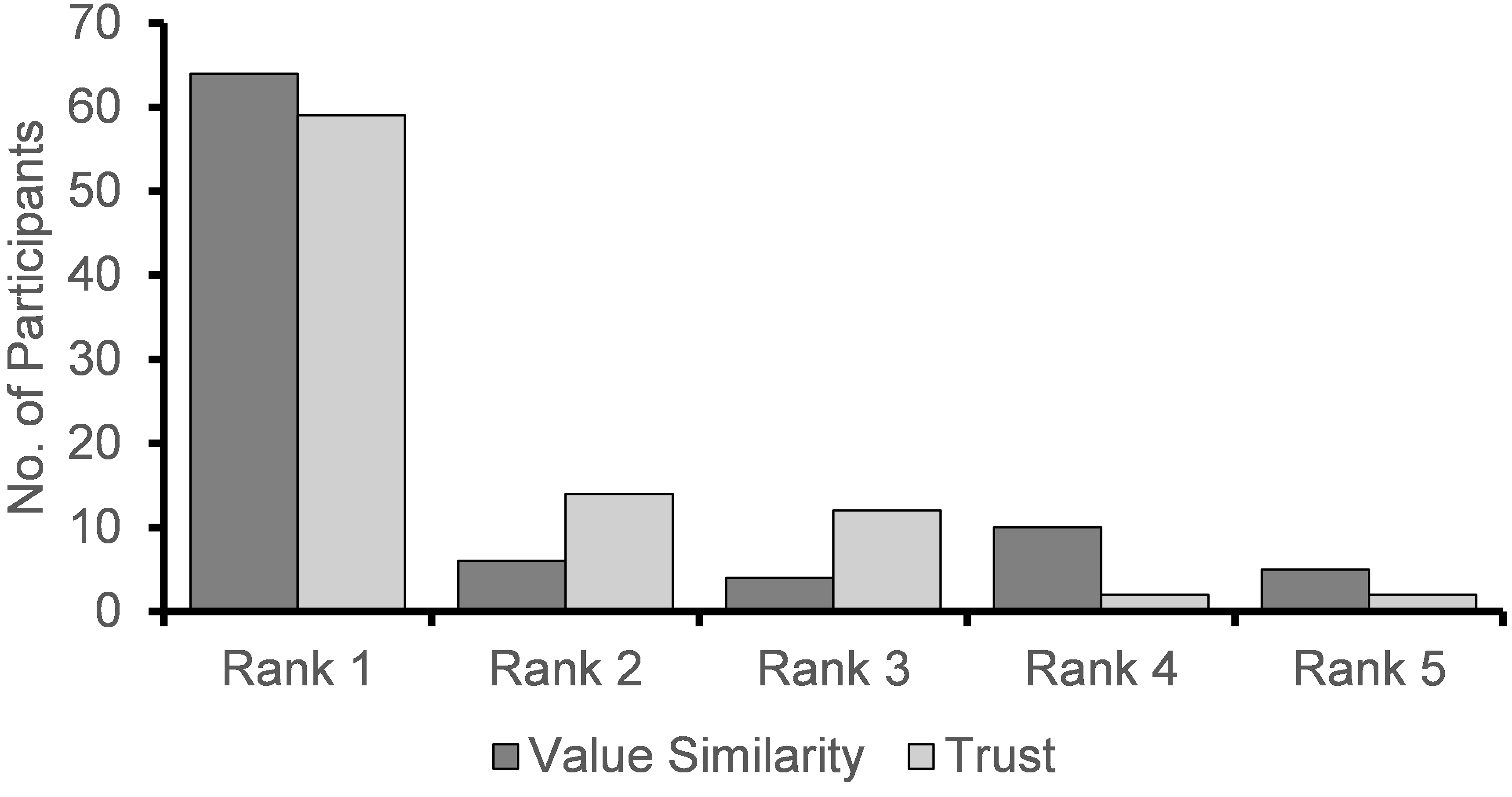}
\caption{This figure represents the number of participants who choose an agent to take inside the building based upon their rank of value similarity and trust.}
\label{fig:vstrustranks}
\end{figure}

We were interested in understanding which agents participants preferred the most. For this, we asked them to choose an agent to take with them inside the building and were asked to explain their reasons for doing so. We analyzed participants responses for selecting an agent. Our results indicate that the participants pick that agent that shares the most similar values. In figure \ref{fig:vstrustranks}, we can observe that more than 72\% of participants chose the agent they ranked highest on value similarity and trust. This gives us another impression that subjective value similarity and trust correlate with each other. Now, we classified participants qualitative explanations into four themes found by thematic analysis \cite{braun2012thematic}. This classification provides insight into the reasons for participant's choices and how it translates to actual behaviour for selecting an agent. The four themes for selecting one agent over others are:
\begin{enumerate}
    \item Common Values - the selected agent had more values in common with the human than the other agents.
    \item Balanced Advice - the selected agent provided more balanced advice to the human participant than the other agents.
    \item Developed Trust - the selected agent's advice/suggestion inclined the participant to trust the agent.
    \item Participant's Belief - the agent was selected based on its advice/suggestion; this decision was neither related to values nor developed trust.
\end{enumerate}
Out of a total of 89 participants only 55 provided an explanation for choosing an agent. Three researchers coded the explanations written by the participants. Each researcher performed the coding with three to four iterations before deciding upon final themes. Inter-coder reliability analysis was performed using Cohen’s kappa to determine agreement and consistency between all coders. There was a near-perfect agreement among all three coders for three dimensions $\kappa$ = .900, (95\% CI, .643 to .937).

Based on our analysis, we found that 42\% of the participants explanations were related to common values between the participant and the agent they chose. This was followed by 23\% for balanced advice given by the agent, 16\% for developed trust and 16\% for belief of the participant. These results shows that in our experiment, VS and balanced advice promoted the intended behaviour of participants to select an agent. For example, P54 said, ‘\textit{He [Agent A] thinks the same way as me so I think he'd back me in my decisions}’ relates to choosing an agent based on the common values. Similarly, P39 said, ‘\textit{I believe agent B thinks 100\% like me and gives me all the trust and responsibility}’ relates to developed trust for the agent. We also came across many responses where participants choose the agent because of balanced advice by them. For example, P44 said, ‘\textit{Agent B shows a balance of risk taking and following protocol to handle a delicate situation}’. Finally, few participants stick to their beliefs for their decision. This can be seen with what P27 reported as ‘\textit{I believe he [Agent B] would be able to help save the hostages and neutralize the threat with non lethal force if possible and lethal if absolutely necessary}’.

\section{Discussion}
In this section, we discuss the results of our study, relating them to prior work and making inferences on how the results can be applied to the design of AI agents. Recall that our main goal was to understand the effect of similarity of human \& agent's values on a human's trust in that agent. Based on our study results, the hypothesis (that the VS between the user and the agent positively affects the trust a user has in that agent) can be partially accepted. We showed that there exists an overall significant effect of VS on trust. Even though our failed manipulations did not interfere with our paper's primary goal, we were intrigued to find out that our manipulations of VS were not successful. In the following section, we discuss possible reasons for our unsuccessful manipulations.

\subsection{Why our manipulations were unsuccessful?}  
If we wish to eventually promote appropriate trust, we should also be able to influence trust. To this end we need to know what factors influence trust, and we need to be able to manipulate these factors in the designs of agents. In this paper we have added to the knowledge on factors that influence trust by showing the relationship with value similarity. However, the manipulation of those factors did not fully succeed in our study. Therefore, it is relevant to examine closer why our manipulations failed and provide some suggestions for how value similarity might be manipulated successfully in the future. 

\begin{figure}[h]
\centering
\includegraphics[width=0.485\textwidth]{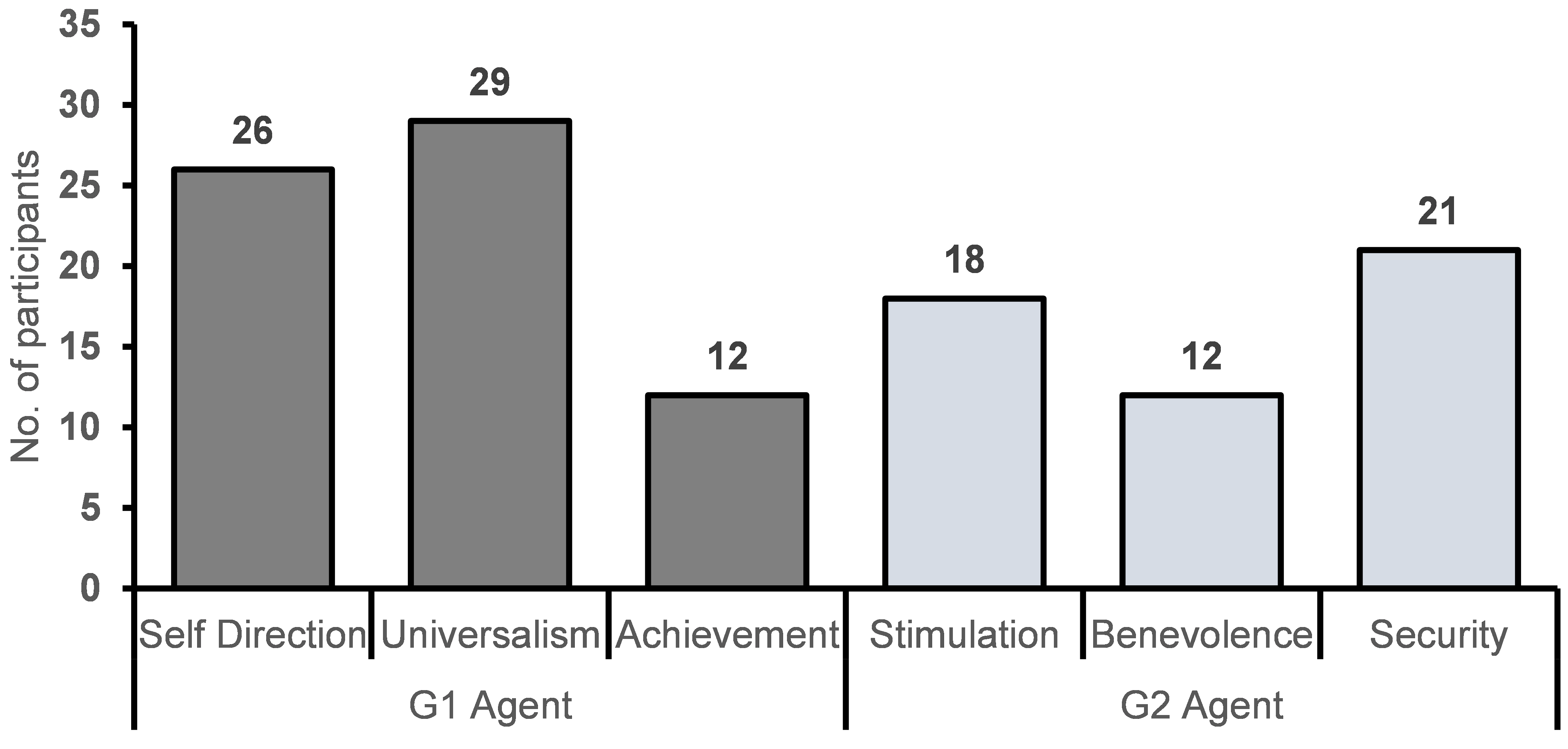}
\caption{Top three most common values in the value profile of the G1 agent (values ranked 1 and 2 of participant) and the G2 agent (values ranked 3 and 4 of the participant). The numbers on the top of the histogram represent how often those values occurred in this agent (G1 or G2) for our participants.
}
\label{fig:valuesagents}
\end{figure}

Regarding our specific agent design, a successful manipulation would have led to the observation that the \textit{`G1'} agent is rated highest for the perceived VS and the \textit{`G5'} agent the least. However, we observed that instead both the \textit{`G2'} and the \textit{`G3'} agent were rated as having more similar values than the \textit{`G1'} agent. To understand why this happened, we examined the actual value profiles of the participants more closely. Consider the case when VS scores of the \textit{`G2'} agent were higher than those of the \textit{`G1'} agent. Observing the participants' specific value profiles for who this occurred could provide us with potential reasons why manipulations were not successful. Figure \ref{fig:valuesagents} provides an overview of the values used in the explanations of the \textit{'G1'} and the \textit{'G2'} agents, and how often those occurred. This figure shows that the values of Self-Direction, Universalism \& Achievement were most prominent for the agent \textit{`G1'} and Stimulation, Benevolence \& Security for the agent \textit{`G2'}, for those participants where \textit{`G2'} scored higher than \textit{`G1'} in value similarity.  
Given that people felt most similar to agents which promoted stimulation, benevolence and security (as opposed to the values of self-direction, universalism and achievement which scored higher in their value profile), we speculate that the choice of scenario might have played a role. The major values for agent \textit{`G1'} - Self Direction and Universalism, were those which participants already possessed but were not so relevant in this context of saving a hostage. On the other hand, for agent \textit{`G2'} - Security and Stimulation were vital because they relate to safety and motivating the participant to save the hostage. It makes intuitive sense that contextual values are of utmost importance especially in those scenarios where there is a risk associated with trusting someone and not all the values are equally salient. However, the value profile survey is general, and not context-dependent.   \begin{tcolorbox}[left=.5em,right=.5em,boxrule=0pt]
Therefore, we speculate that when designing value profiles for artificial agents, one should not just take into account general value profiles, but also note which contextual values are most important as also echoed by Liscio et al. \cite{liscio2021axies}.
\end{tcolorbox}

Another potential reason for our failed manipulation could be that a discrepancy existed regarding values of the agent in how they were perceived by some of the participants and how they were intended. By perceived values we mean that the value laden explanations that agents provided were sometimes interpreted as promoting different values than for which they were written. As explained in section `Scenario and agent explanation', it took three iterations for each explanation to be finalized, which indicates how quickly disagreements about underlying values of explanations can occur. We speculate that this discrepancy is a possible reason for our failed manipulation and resound with Wang et al. \cite{wang2019designing} that designing agent explanations that can be consistently interpreted by humans is still an open research area. Secondly, consistency in value preferences from humans is debated, and people could just show inconsistencies as mentioned by Boyd et al. \cite{boyd2015values}.

\subsection{Trust in AI systems}
In our user-study, agents provided their suggestions based on value-based reasoning using VS. With the use of value-based reasoning, an agent includes the representation for human values and can provide reasoning using human values to make decisions. Winikoff argue that a computational model of relevant human values can be used to provide higher level, human-centered explanations of decisions by AI agents. This means that agents could use value-based reasoning when trying to influence trust \cite{winikoff2017towards}. In synopsis, given a bunch of random generated agents, humans would trust those align with their subjective values. The reported correlation can also comes from human’s consistent value judgment about the suggestions and scenarios.

Value Similarity is not the only thing that influences trust; many other factors can influence trust as well. Three of the main aspects of trust are benevolence, willingness and competence. Value similarity could be seen as a part of benevolence, or even willingness. However, competence is less related to values \cite{wojciszke1997parallels}. We did not focus upon this factor because we provided all our agents a ground truth \textit{i.e.} prior common knowledge. Instead, we focused upon benevolence and willingness as other two factors of trust affected by VS in accordance with Gulati et al. \cite{gulati2019design}. Based on our results, both these factors were moderately positively correlated with VS. This implies that if we wish to understand how humans trust systems we need to look beyond trust as being influenced only by the system's reliability. Rather, we also need to consider trust in benevolence and willingness and understand how these are influenced by aspects such as value similarity \cite{cohen2019trusted}. 

\subsection{Limitations \& Future Work}
We investigated the effect of VS on trust with a risk-taking scenario of saving a hostage. We chose this scenario to gain a deeper understanding of how participants trust an agent with most to least VS. However, we believe that further evaluation with more real-life examples would provide additional insights on participant's trust. Additionally, although we cross-examined the participant value profile with their responses to the VS questionnaire, we did not focus on their understanding of value laden explanations. We posit that examining the perception of the values could have provided a more subtle effect of our manipulations. We see this as an opportunity to further extend our work into understanding the beliefs and perceptions of the participants for agents with varying VS. Also, future work could extend the proposed method to multiple scenarios with different context information. Additionally, crowdsourcing could be another way to generate explanations instead of pre-designing by experts or experimenters, especially in translating abstract values to specific descriptions or behaviors.  Finally, as explained in the previous section, we would like to study the potential effect of culture on our findings. 

\section{Conclusion}
Our study shows that value similarity between an agent and a human is positively related to how much that human trusts the agent. Based on this finding, we would encourage designers of explanation and feedback-giving agents to create agents that outline human values. An agent with similar values to the human will be trusted more which can be very important in any risk-taking scenario. Although a system without value-based reasoning may be easier to develop, the benefits of including VS are worth it, especially in trust-critical situations. 

\appendix
\section*{Supplementary}
The raw data set of this study along with the processed data files are available at \url{https://doi.org/10.4121/14518380}.
\begin{acks}
Special thanks go to Ilir Kola and Sandy Manolios who helped in designing the explanations for the user study, and anonymous reviewers who provided useful suggestions for incorporation in this work. 
\end{acks}

\bibliographystyle{ACM-Reference-Format}
\balance
\bibliography{acmart}


\begin{thebibliography}{36}


\ifx \showCODEN    \undefined \def \showCODEN     #1{\unskip}     \fi
\ifx \showDOI      \undefined \def \showDOI       #1{#1}\fi
\ifx \showISBNx    \undefined \def \showISBNx     #1{\unskip}     \fi
\ifx \showISBNxiii \undefined \def \showISBNxiii  #1{\unskip}     \fi
\ifx \showISSN     \undefined \def \showISSN      #1{\unskip}     \fi
\ifx \showLCCN     \undefined \def \showLCCN      #1{\unskip}     \fi
\ifx \shownote     \undefined \def \shownote      #1{#1}          \fi
\ifx \showarticletitle \undefined \def \showarticletitle #1{#1}   \fi
\ifx \showURL      \undefined \def \showURL       {\relax}        \fi
\providecommand\bibfield[2]{#2}
\providecommand\bibinfo[2]{#2}
\providecommand\natexlab[1]{#1}
\providecommand\showeprint[2][]{arXiv:#2}

\bibitem[\protect\citeauthoryear{Boyd, Wilson, Pennebaker, Kosinski, Stillwell,
  and Mihalcea}{Boyd et~al\mbox{.}}{2015}]%
        {boyd2015values}
\bibfield{author}{\bibinfo{person}{Ryan Boyd}, \bibinfo{person}{Steven Wilson},
  \bibinfo{person}{James Pennebaker}, \bibinfo{person}{Michal Kosinski},
  \bibinfo{person}{David Stillwell}, {and} \bibinfo{person}{Rada Mihalcea}.}
  \bibinfo{year}{2015}\natexlab{}.
\newblock \showarticletitle{Values in words: Using language to evaluate and
  understand personal values}. In \bibinfo{booktitle}{\emph{Proceedings of the
  International AAAI Conference on Web and Social Media}},
  Vol.~\bibinfo{volume}{9}.
\newblock


\bibitem[\protect\citeauthoryear{Braun and Clarke}{Braun and Clarke}{2020}]%
        {braun2012thematic}
\bibfield{author}{\bibinfo{person}{Virginia Braun} {and}
  \bibinfo{person}{Victoria Clarke}.} \bibinfo{year}{2020}\natexlab{}.
\newblock \showarticletitle{One size fits all? What counts as quality practice
  in (reflexive) thematic analysis?}
\newblock \bibinfo{journal}{\emph{Qualitative research in psychology}}
  (\bibinfo{year}{2020}), \bibinfo{pages}{1--25}.
\newblock


\bibitem[\protect\citeauthoryear{Chavaillaz, Wastell, and Sauer}{Chavaillaz
  et~al\mbox{.}}{2016}]%
        {chavaillaz2016system}
\bibfield{author}{\bibinfo{person}{Alain Chavaillaz}, \bibinfo{person}{David
  Wastell}, {and} \bibinfo{person}{J{\"u}rgen Sauer}.}
  \bibinfo{year}{2016}\natexlab{}.
\newblock \showarticletitle{System reliability, performance and trust in
  adaptable automation}.
\newblock \bibinfo{journal}{\emph{Applied Ergonomics}}  \bibinfo{volume}{52}
  (\bibinfo{year}{2016}), \bibinfo{pages}{333--342}.
\newblock


\bibitem[\protect\citeauthoryear{Chhogyal, Nayak, Ghose, and Dam}{Chhogyal
  et~al\mbox{.}}{2019}]%
        {ijcai2019-28}
\bibfield{author}{\bibinfo{person}{Kinzang Chhogyal}, \bibinfo{person}{Abhaya
  Nayak}, \bibinfo{person}{Aditya Ghose}, {and} \bibinfo{person}{Hoa~K. Dam}.}
  \bibinfo{year}{2019}\natexlab{}.
\newblock \showarticletitle{A Value-based Trust Assessment Model for
  Multi-agent Systems}. In \bibinfo{booktitle}{\emph{Proceedings of the
  Twenty-Eighth International Joint Conference on Artificial Intelligence,
  {IJCAI-19}}}. \bibinfo{publisher}{International Joint Conferences on
  Artificial Intelligence Organization}, \bibinfo{pages}{194--200}.
\newblock
\urldef\tempurl%
\url{https://doi.org/10.24963/ijcai.2019/28}
\showDOI{\tempurl}


\bibitem[\protect\citeauthoryear{Cohen, Schaekermann, Liu, and Cormier}{Cohen
  et~al\mbox{.}}{2019}]%
        {cohen2019trusted}
\bibfield{author}{\bibinfo{person}{Robin Cohen}, \bibinfo{person}{Mike
  Schaekermann}, \bibinfo{person}{Sihao Liu}, {and} \bibinfo{person}{Michael
  Cormier}.} \bibinfo{year}{2019}\natexlab{}.
\newblock \showarticletitle{Trusted Ai and the Contribution of Trust Modeling
  in Multiagent Systems}. In \bibinfo{booktitle}{\emph{Proceedings of the 18th
  International Conference on Autonomous Agents and MultiAgent Systems}}.
  International Foundation for Autonomous Agents and Multiagent Systems,
  \bibinfo{pages}{1644--1648}.
\newblock


\bibitem[\protect\citeauthoryear{Cranefield, Winikoff, Dignum, and
  Dignum}{Cranefield et~al\mbox{.}}{2017}]%
        {cranefield2017no}
\bibfield{author}{\bibinfo{person}{Stephen Cranefield},
  \bibinfo{person}{Michael Winikoff}, \bibinfo{person}{Virginia Dignum}, {and}
  \bibinfo{person}{Frank Dignum}.} \bibinfo{year}{2017}\natexlab{}.
\newblock \showarticletitle{No Pizza for You: Value-based Plan Selection in BDI
  Agents.}. In \bibinfo{booktitle}{\emph{IJCAI}}. \bibinfo{pages}{178--184}.
\newblock


\bibitem[\protect\citeauthoryear{Cruciani, Moretti, and Pellizzari}{Cruciani
  et~al\mbox{.}}{2017}]%
        {cruciani2017dynamic}
\bibfield{author}{\bibinfo{person}{Caterina Cruciani}, \bibinfo{person}{Anna
  Moretti}, {and} \bibinfo{person}{Paolo Pellizzari}.}
  \bibinfo{year}{2017}\natexlab{}.
\newblock \showarticletitle{Dynamic Patterns in Similarity-based Cooperation:
  An Agent-based Investigation}.
\newblock \bibinfo{journal}{\emph{Journal of Economic Interaction and
  Coordination}} \bibinfo{volume}{12}, \bibinfo{number}{1}
  (\bibinfo{year}{2017}), \bibinfo{pages}{121--141}.
\newblock


\bibitem[\protect\citeauthoryear{Dechesne, Di~Tosto, Dignum, and
  Dignum}{Dechesne et~al\mbox{.}}{2013}]%
        {dechesne2013no}
\bibfield{author}{\bibinfo{person}{Francien Dechesne}, \bibinfo{person}{Gennaro
  Di~Tosto}, \bibinfo{person}{Virginia Dignum}, {and} \bibinfo{person}{Frank
  Dignum}.} \bibinfo{year}{2013}\natexlab{}.
\newblock \showarticletitle{No Smoking Here: Values, Norms and Culture in
  Multi-agent Systems}.
\newblock \bibinfo{journal}{\emph{Artificial intelligence and law}}
  \bibinfo{volume}{21}, \bibinfo{number}{1} (\bibinfo{year}{2013}),
  \bibinfo{pages}{79--107}.
\newblock


\bibitem[\protect\citeauthoryear{Faul, Erdfelder, Lang, and Buchner}{Faul
  et~al\mbox{.}}{2007}]%
        {faul2007g}
\bibfield{author}{\bibinfo{person}{Franz Faul}, \bibinfo{person}{Edgar
  Erdfelder}, \bibinfo{person}{Albert-Georg Lang}, {and} \bibinfo{person}{Axel
  Buchner}.} \bibinfo{year}{2007}\natexlab{}.
\newblock \showarticletitle{G* Power 3: A Flexible Statistical Power Analysis
  Program for the Social, Behavioral, and Biomedical Sciences}.
\newblock \bibinfo{journal}{\emph{Behavior research methods}}
  \bibinfo{volume}{39}, \bibinfo{number}{2} (\bibinfo{year}{2007}),
  \bibinfo{pages}{175--191}.
\newblock


\bibitem[\protect\citeauthoryear{Friedman, Kahn, and Borning}{Friedman
  et~al\mbox{.}}{2008}]%
        {friedman2008value}
\bibfield{author}{\bibinfo{person}{Batya Friedman}, \bibinfo{person}{Peter~H
  Kahn}, {and} \bibinfo{person}{Alan Borning}.}
  \bibinfo{year}{2008}\natexlab{}.
\newblock \showarticletitle{Value Sensitive Design and Information Systems}.
\newblock \bibinfo{journal}{\emph{The handbook of information and computer
  ethics}} (\bibinfo{year}{2008}), \bibinfo{pages}{69--101}.
\newblock


\bibitem[\protect\citeauthoryear{Glikson and Woolley}{Glikson and
  Woolley}{2020}]%
        {glikson2020human}
\bibfield{author}{\bibinfo{person}{Ella Glikson} {and}
  \bibinfo{person}{Anita~Williams Woolley}.} \bibinfo{year}{2020}\natexlab{}.
\newblock \showarticletitle{Human Trust in Artificial Intelligence: Review of
  Empirical Research}.
\newblock \bibinfo{journal}{\emph{Academy of Management Annals}}
  \bibinfo{number}{14, 2} (\bibinfo{year}{2020}).
\newblock


\bibitem[\protect\citeauthoryear{Gulati, Sousa, and Lamas}{Gulati
  et~al\mbox{.}}{2019}]%
        {gulati2019design}
\bibfield{author}{\bibinfo{person}{Siddharth Gulati}, \bibinfo{person}{Sonia
  Sousa}, {and} \bibinfo{person}{David Lamas}.}
  \bibinfo{year}{2019}\natexlab{}.
\newblock \showarticletitle{Design, Development and Evaluation of a
  Human-Computer Trust Scale}.
\newblock \bibinfo{journal}{\emph{Behaviour \&; Information Technology}}
  \bibinfo{volume}{38}, \bibinfo{number}{10} (\bibinfo{year}{2019}),
  \bibinfo{pages}{1004--1015}.
\newblock


\bibitem[\protect\citeauthoryear{Guru}{Guru}{2016}]%
        {guru_2016}
\bibfield{author}{\bibinfo{person}{Banavar Guru}.}
  \bibinfo{year}{2016}\natexlab{}.
\newblock \showarticletitle{What {It} {Will} {Take} for {Us} to {Trust} {AI}}.
\newblock \bibinfo{journal}{\emph{Harvard Business Review}}
  (\bibinfo{date}{Nov.} \bibinfo{year}{2016}).
\newblock
\showISSN{0017-8012}
\urldef\tempurl%
\url{https://hbr.org/2016/11/what-it-will-take-for-us-to-trust-ai}
\showURL{%
\tempurl}


\bibitem[\protect\citeauthoryear{Liscio, van~der Meer, Siebert, Jonker, Mouter,
  and Murukannaiah}{Liscio et~al\mbox{.}}{2021}]%
        {liscio2021axies}
\bibfield{author}{\bibinfo{person}{Enrico Liscio}, \bibinfo{person}{Michiel
  van~der Meer}, \bibinfo{person}{Luciano~C Siebert},
  \bibinfo{person}{Catholijn~M Jonker}, \bibinfo{person}{Niek Mouter}, {and}
  \bibinfo{person}{Pradeep~K Murukannaiah}.} \bibinfo{year}{2021}\natexlab{}.
\newblock \showarticletitle{Axies: Identifying and Evaluating Context-Specific
  Values}. In \bibinfo{booktitle}{\emph{Proceedings of the 20th International
  Conference on Autonomous Agents and MultiAgent Systems}}.
  \bibinfo{pages}{799--808}.
\newblock


\bibitem[\protect\citeauthoryear{Mercuur, Dignum, and Jonker}{Mercuur
  et~al\mbox{.}}{2019}]%
        {mercuur2019value}
\bibfield{author}{\bibinfo{person}{Rijk Mercuur}, \bibinfo{person}{Virginia
  Dignum}, {and} \bibinfo{person}{Catholijn Jonker}.}
  \bibinfo{year}{2019}\natexlab{}.
\newblock \showarticletitle{The Value of Values and Norms in Social
  Simulation}.
\newblock \bibinfo{journal}{\emph{Journal of Artificial Societies and Social
  Simulation}} \bibinfo{volume}{22}, \bibinfo{number}{1}
  (\bibinfo{year}{2019}).
\newblock


\bibitem[\protect\citeauthoryear{Ogawa, Park, and Umemuro}{Ogawa
  et~al\mbox{.}}{2019}]%
        {ogawa2019humans}
\bibfield{author}{\bibinfo{person}{Rui Ogawa}, \bibinfo{person}{Sung Park},
  {and} \bibinfo{person}{Hiroyuki Umemuro}.} \bibinfo{year}{2019}\natexlab{}.
\newblock \showarticletitle{How Humans Develop Trust in Communication Robots: A
  Phased Model Based on Interpersonal Trust}. In \bibinfo{booktitle}{\emph{2019
  14th ACM/IEEE International Conference on Human-Robot Interaction (HRI)}}.
  IEEE, \bibinfo{pages}{606--607}.
\newblock


\bibitem[\protect\citeauthoryear{Parasuraman and Riley}{Parasuraman and
  Riley}{1997}]%
        {parasuraman1997humans}
\bibfield{author}{\bibinfo{person}{Raja Parasuraman} {and}
  \bibinfo{person}{Victor Riley}.} \bibinfo{year}{1997}\natexlab{}.
\newblock \showarticletitle{Humans and automation: Use, misuse, disuse, abuse}.
\newblock \bibinfo{journal}{\emph{Human factors}} \bibinfo{volume}{39},
  \bibinfo{number}{2} (\bibinfo{year}{1997}), \bibinfo{pages}{230--253}.
\newblock


\bibitem[\protect\citeauthoryear{Partovi, Zukerman, Zhan, Hamacher, and
  Hohwy}{Partovi et~al\mbox{.}}{2019}]%
        {partovi2019relationship}
\bibfield{author}{\bibinfo{person}{Andisheh Partovi}, \bibinfo{person}{Ingrid
  Zukerman}, \bibinfo{person}{Kai Zhan}, \bibinfo{person}{Nora Hamacher}, {and}
  \bibinfo{person}{Jakob Hohwy}.} \bibinfo{year}{2019}\natexlab{}.
\newblock \showarticletitle{Relationship between Device Performance, Trust and
  User Behaviour in a Care-taking Scenario}. In
  \bibinfo{booktitle}{\emph{Proceedings of the 27th ACM Conference on User
  Modeling, Adaptation and Personalization}}. \bibinfo{pages}{61--69}.
\newblock


\bibitem[\protect\citeauthoryear{Pe{\~n}a and Slate}{Pe{\~n}a and
  Slate}{2006}]%
        {pena2006global}
\bibfield{author}{\bibinfo{person}{Edsel~A Pe{\~n}a} {and}
  \bibinfo{person}{Elizabeth~H Slate}.} \bibinfo{year}{2006}\natexlab{}.
\newblock \showarticletitle{Global Validation of Linear Model Assumptions}.
\newblock \bibinfo{journal}{\emph{J. Amer. Statist. Assoc.}}
  \bibinfo{volume}{101}, \bibinfo{number}{473} (\bibinfo{year}{2006}),
  \bibinfo{pages}{341--354}.
\newblock


\bibitem[\protect\citeauthoryear{Rajagopalachari}{Rajagopalachari}{1970}]%
        {rajagopalachari1970mahabharata}
\bibfield{author}{\bibinfo{person}{Chakravarti Rajagopalachari}.}
  \bibinfo{year}{1970}\natexlab{}.
\newblock \bibinfo{booktitle}{\emph{Mahabharata}}. Vol.~\bibinfo{volume}{1}.
\newblock \bibinfo{publisher}{Diamond Pocket Books (P) Ltd.}
\newblock


\bibitem[\protect\citeauthoryear{Ratner}{Ratner}{2009}]%
        {ratner2009correlation}
\bibfield{author}{\bibinfo{person}{Bruce Ratner}.}
  \bibinfo{year}{2009}\natexlab{}.
\newblock \showarticletitle{The correlation coefficient: Its values range
  between+ 1/- 1, or do they?}
\newblock \bibinfo{journal}{\emph{Journal of targeting, measurement and
  analysis for marketing}} \bibinfo{volume}{17}, \bibinfo{number}{2}
  (\bibinfo{year}{2009}), \bibinfo{pages}{139--142}.
\newblock


\bibitem[\protect\citeauthoryear{Roff and Danks}{Roff and Danks}{2018}]%
        {roff2018trust}
\bibfield{author}{\bibinfo{person}{Heather~M Roff} {and} \bibinfo{person}{David
  Danks}.} \bibinfo{year}{2018}\natexlab{}.
\newblock \showarticletitle{“Trust but Verify”: The Difficulty of Trusting
  Autonomous Weapons Systems}.
\newblock \bibinfo{journal}{\emph{Journal of Military Ethics}}
  \bibinfo{volume}{17}, \bibinfo{number}{1} (\bibinfo{year}{2018}),
  \bibinfo{pages}{2--20}.
\newblock


\bibitem[\protect\citeauthoryear{Ryan}{Ryan}{2020}]%
        {ryan2020ai}
\bibfield{author}{\bibinfo{person}{Mark Ryan}.}
  \bibinfo{year}{2020}\natexlab{}.
\newblock \showarticletitle{In AI We Trust: Ethics, Artificial Intelligence,
  and Reliability}.
\newblock \bibinfo{journal}{\emph{Science and Engineering Ethics}}
  (\bibinfo{year}{2020}), \bibinfo{pages}{1--19}.
\newblock


\bibitem[\protect\citeauthoryear{Schwartz}{Schwartz}{2012}]%
        {schwartz2012overview}
\bibfield{author}{\bibinfo{person}{Shalom~H Schwartz}.}
  \bibinfo{year}{2012}\natexlab{}.
\newblock \showarticletitle{An Overview of the Schwartz Theory of Basic
  Values}.
\newblock \bibinfo{journal}{\emph{Online readings in Psychology and Culture}}
  \bibinfo{volume}{2}, \bibinfo{number}{1} (\bibinfo{year}{2012}),
  \bibinfo{pages}{2307--0919}.
\newblock


\bibitem[\protect\citeauthoryear{Shapiro and Shachter}{Shapiro and
  Shachter}{2002}]%
        {shapiro2002user}
\bibfield{author}{\bibinfo{person}{Daniel Shapiro} {and} \bibinfo{person}{Ross
  Shachter}.} \bibinfo{year}{2002}\natexlab{}.
\newblock \showarticletitle{User-agent Value Alignment}. In
  \bibinfo{booktitle}{\emph{Proc. of The 18th Nat. Conf. on Artif. Intell.
  AAAI}}.
\newblock


\bibitem[\protect\citeauthoryear{Shvo, Buhmann, and Kapadia}{Shvo
  et~al\mbox{.}}{2019}]%
        {shvo2019towards}
\bibfield{author}{\bibinfo{person}{Maayan Shvo}, \bibinfo{person}{Jakob
  Buhmann}, {and} \bibinfo{person}{Mubbasir Kapadia}.}
  \bibinfo{year}{2019}\natexlab{}.
\newblock \showarticletitle{Towards Modeling the Interplay of Personality,
  Motivation, Emotion, and Mood in Social Agents}. In
  \bibinfo{booktitle}{\emph{Proceedings of the 18th International Conference on
  Autonomous Agents and MultiAgent Systems}}. \bibinfo{pages}{2195--2197}.
\newblock


\bibitem[\protect\citeauthoryear{Siegrist, Cvetkovich, and Roth}{Siegrist
  et~al\mbox{.}}{2000}]%
        {siegrist2000salient}
\bibfield{author}{\bibinfo{person}{Michael Siegrist}, \bibinfo{person}{George
  Cvetkovich}, {and} \bibinfo{person}{Claudia Roth}.}
  \bibinfo{year}{2000}\natexlab{}.
\newblock \showarticletitle{Salient Value Similarity, Social Trust, and
  Risk/Benefit Perception}.
\newblock \bibinfo{journal}{\emph{Risk analysis}} \bibinfo{volume}{20},
  \bibinfo{number}{3} (\bibinfo{year}{2000}), \bibinfo{pages}{353--362}.
\newblock


\bibitem[\protect\citeauthoryear{Sitkin and Roth}{Sitkin and Roth}{1993}]%
        {sitkin1993explaining}
\bibfield{author}{\bibinfo{person}{Sim~B Sitkin} {and} \bibinfo{person}{Nancy~L
  Roth}.} \bibinfo{year}{1993}\natexlab{}.
\newblock \showarticletitle{Explaining the limited effectiveness of legalistic
  “remedies” for trust/distrust}.
\newblock \bibinfo{journal}{\emph{Organization science}} \bibinfo{volume}{4},
  \bibinfo{number}{3} (\bibinfo{year}{1993}), \bibinfo{pages}{367--392}.
\newblock


\bibitem[\protect\citeauthoryear{Strack and Gennerich}{Strack and
  Gennerich}{2011}]%
        {strack2011personal}
\bibfield{author}{\bibinfo{person}{Micha Strack} {and} \bibinfo{person}{Carsten
  Gennerich}.} \bibinfo{year}{2011}\natexlab{}.
\newblock \showarticletitle{Personal and Situational Values Predict Ethical
  Reasoning}.
\newblock \bibinfo{journal}{\emph{Europe’s Journal of Psychology}}
  \bibinfo{volume}{7}, \bibinfo{number}{3} (\bibinfo{year}{2011}),
  \bibinfo{pages}{419--442}.
\newblock


\bibitem[\protect\citeauthoryear{Tolmeijer, Weiss, Hanheide, Lindner, Powers,
  Dixon, and Tielman}{Tolmeijer et~al\mbox{.}}{2020}]%
        {tolmeijer2020taxonomy}
\bibfield{author}{\bibinfo{person}{Suzanne Tolmeijer}, \bibinfo{person}{Astrid
  Weiss}, \bibinfo{person}{Marc Hanheide}, \bibinfo{person}{Felix Lindner},
  \bibinfo{person}{Thomas~M Powers}, \bibinfo{person}{Clare Dixon}, {and}
  \bibinfo{person}{Myrthe~L Tielman}.} \bibinfo{year}{2020}\natexlab{}.
\newblock \showarticletitle{Taxonomy of Trust-Relevant Failures and Mitigation
  Strategies}. In \bibinfo{booktitle}{\emph{Proceedings of the 2020 ACM/IEEE
  International Conference on Human-Robot Interaction}}.
  \bibinfo{pages}{3--12}.
\newblock


\bibitem[\protect\citeauthoryear{Van~Riemsdijk, Jonker, and
  Lesser}{Van~Riemsdijk et~al\mbox{.}}{2015}]%
        {van2015creating}
\bibfield{author}{\bibinfo{person}{M~Birna Van~Riemsdijk},
  \bibinfo{person}{Catholijn~M Jonker}, {and} \bibinfo{person}{Victor Lesser}.}
  \bibinfo{year}{2015}\natexlab{}.
\newblock \showarticletitle{Creating Socially Adaptive Electronic Partners:
  Interaction, Reasoning and Ethical Challenges}. In
  \bibinfo{booktitle}{\emph{Proceedings of the 2015 international conference on
  autonomous agents and multiagent systems}}. \bibinfo{pages}{1201--1206}.
\newblock


\bibitem[\protect\citeauthoryear{Vaske, Absher, and Bright}{Vaske
  et~al\mbox{.}}{2007}]%
        {vaske2007salient}
\bibfield{author}{\bibinfo{person}{Jerry~J Vaske}, \bibinfo{person}{James~D
  Absher}, {and} \bibinfo{person}{Alan~D Bright}.}
  \bibinfo{year}{2007}\natexlab{}.
\newblock \showarticletitle{Salient Value Similarity, Social Trust and
  Attitudes Toward Wildland Fire Management Strategies}.
\newblock \bibinfo{journal}{\emph{Human Ecology Review}}
  (\bibinfo{year}{2007}), \bibinfo{pages}{223--232}.
\newblock


\bibitem[\protect\citeauthoryear{Wang, Yang, Abdul, and Lim}{Wang
  et~al\mbox{.}}{2019}]%
        {wang2019designing}
\bibfield{author}{\bibinfo{person}{Danding Wang}, \bibinfo{person}{Qian Yang},
  \bibinfo{person}{Ashraf Abdul}, {and} \bibinfo{person}{Brian~Y Lim}.}
  \bibinfo{year}{2019}\natexlab{}.
\newblock \showarticletitle{Designing theory-driven user-centric explainable
  AI}. In \bibinfo{booktitle}{\emph{Proceedings of the 2019 CHI conference on
  human factors in computing systems}}. \bibinfo{pages}{1--15}.
\newblock


\bibitem[\protect\citeauthoryear{Wang, Pynadath, and Hill}{Wang
  et~al\mbox{.}}{2016}]%
        {wang2016impact}
\bibfield{author}{\bibinfo{person}{Ning Wang}, \bibinfo{person}{David~V
  Pynadath}, {and} \bibinfo{person}{Susan~G Hill}.}
  \bibinfo{year}{2016}\natexlab{}.
\newblock \showarticletitle{The impact of pomdp-generated explanations on trust
  and performance in human-robot teams}. In
  \bibinfo{booktitle}{\emph{Proceedings of the 2016 international conference on
  autonomous agents \&; multiagent systems}}. \bibinfo{pages}{997--1005}.
\newblock


\bibitem[\protect\citeauthoryear{Winikoff}{Winikoff}{2017}]%
        {winikoff2017towards}
\bibfield{author}{\bibinfo{person}{Michael Winikoff}.}
  \bibinfo{year}{2017}\natexlab{}.
\newblock \showarticletitle{Towards Trusting Autonomous Systems}. In
  \bibinfo{booktitle}{\emph{International Workshop on Engineering Multi-Agent
  Systems}}. Springer, \bibinfo{pages}{3--20}.
\newblock


\bibitem[\protect\citeauthoryear{Wojciszke}{Wojciszke}{1997}]%
        {wojciszke1997parallels}
\bibfield{author}{\bibinfo{person}{Bogdan Wojciszke}.}
  \bibinfo{year}{1997}\natexlab{}.
\newblock \showarticletitle{Parallels Between Competence-versus
  Morality-related Traits and Individualistic Versus Collectivistic Values}.
\newblock \bibinfo{journal}{\emph{European Journal of Social Psychology}}
  \bibinfo{volume}{27}, \bibinfo{number}{3} (\bibinfo{year}{1997}),
  \bibinfo{pages}{245--256}.
\newblock


\end{thebibliography}


\end{document}